\documentclass[]{ceurart}

\usepackage{listings}
\usepackage{booktabs}
\usepackage{hyperref}

\usepackage{float}

\begin{document}

\copyrightyear{2026}
\copyrightclause{Copyright for this paper by its authors.
  Use permitted under Creative Commons License Attribution 4.0
  International (CC BY 4.0).}

\conference{}

\title{eCREAM-MedCorpus: A Large-Scale Corpus of  Clinical Notes for Italian}

\tnotemark[1]
\tnotetext[1]{Dataset and code are available at https://huggingface.co/datasets/NLP-FBK/ecream-emergency-department-notes.}

\author[1]{Tiziano Labruna}[%
  orcid=0000-0001-7713-7679,
  email=tlabruna@fbk.eu,
]\cormark[1]
\author[2]{Guido Bertolini}[%
  orcid=0000-0002-3101-4259,
  email=guido.bertolini@marionegri.it,
]
\author[1,3]{Pietro Ferrazzi}[%
  orcid=0009-0007-6223-9112,
  email=pferrazzi@fbk.eu,
]
\author[3]{Bernardo Magnini}[%
  orcid=0000-0002-0740-5778,
  email=magnini@fbk.eu,
]

\address[1]{Fondazione Bruno Kessler, Trento, Italy}
\address[2]{Istituto di Ricerche Farmacologiche Mario Negri IRCCS, Milan, Italy}
\address[3]{University of Padua, Padua, Italy}

\cortext[1]{Corresponding author.}

\begin{abstract}
We present eCREAM-MedCorpus, a new and unique large-scale dataset of clinical notes produced in Emergency Departments of Italian hospitals. The corpus, in its current version, is composed of approximately 4 million clinical notes fully anonymized, covering diverse phases of patient care during the stay in the emergency department. In addition, a subset of about six thousand notes has been manually annotated by clinical experts through a structured Case Report Form (CRF) containing 132 items relevant for two patient situations in emergency departments, dyspnea and loss of consciousness. Items may assume numerical values (e.g., for blood saturation), categorical (e.g., for level of consciousness ), binary (e.g., for presence of traumas), and mixed value types.
The annotation process involved multiple clinicians and underwent iterative revision to resolve ambiguities in item formulation, resulting in a richly structured (although high imbalanced) resource. The dataset aims to fill a relevant gap of data able to support both the development and the use of Large Language Models in concrete medical applications. 
We describe the data collection protocol, the on-site anonymisation pipeline, corpus statistics, and the annotation scheme.
Finally, we propose CRF-filling as a novel structured information extraction benchmark, and provide zero-shot baseline resulting from Gemma-27B and MedGemma-27B. To the best of our knowledge, eCREAM-MedCorpus is the largest freely available dataset of clinical notes existing for the Italian language.
\end{abstract}

\begin{keywords}
  clinical NLP \sep
  Italian language resources \sep
  information extraction \sep
  CRF filling \sep
  dataset
\end{keywords}

\maketitle

\section{Introduction}

Clinical notes are free text documents produced in hospitals that contain a wealth of information about patients, such as their medical histories, diagnoses, treatments, and outcomes. Yet, clinical notes remain largely inaccessible for research purposes due to a variety of barriers, including regulatory and technical ones. From a regulatory standpoint, health data is considered a special category of sensitive data under frameworks such as the GDPR, and requires strict governance that limits research use. On the technical side, clinical notes are inherently unstructured, written in natural language that varies widely across clinicians, departments, and institutions. This makes automated extraction of meaningful information challenging. To further complicate matters, health data is scattered across isolated systems (hospitals, laboratories, pharmacies, etc.) and relies on inconsistent coding standards and outdated infrastructure. As a consequence,  there is a general scarcity of  real clinical data, which is often addressed by using scientific medical data (e.g.,  articles available from PubMed). However, for specific applications such as the automatic compilation of digital health records, data from scientific papers cannot replace real clinical data. Furthermore, the lack of  clinical data is a particularly pressing issue for languages other than English, such as Italian.

Aiming at reducing the scarcity of clinical data available for research purposes, we present eCREAM-MedCorpus, as a new and unique collection of clinical notes produced in the Emergency Departments of Italian hospitals. The data collection and annotation have been carried out as one of the main activities within eCREAM (enabling Clinical Research in Emergency and Acute care Medicine), a five-year project funded by the Horizon Europe programme (grant agreement no. 101057726). It brings together 11 partners from eight European countries: France, Greece, England, Italy, Poland, Slovakia, Slovenia, and Switzerland. The project’s primary goal is to extract high-quality data suitable for research purposes directly from the electronic health records (EHRs) used in emergency departments (EDs) by exploiting large language models (LLMs). EDs occupy an important position in the healthcare system as they are the primary point of entry for acute and unplanned care. Yet emergency medicine has historically lagged behind other specialities in terms of evidence-based research, largely due to the fragmented, heterogeneous, and time-pressured nature of its practices, which makes traditional manual data collection rarely feasible.

A further core commitment of eCREAM is the FAIRification of the data it produces (making curated datasets Findable, Accessible, Interoperable, and Reusable) in full compliance with European data protection regulations. In this context, the project is building one of the largest multilingual, multicentre corpora of ED clinical notes in Europe across five languages, with expert-annotated subsets designed to support cross-lingual NLP research. The Italian subcorpus, collected so far by the first two participating Italian EDs, forms the basis for the eCREAM-MedCorpus resource presented in this paper and will subsequently be expanded with data from the other Italian participating hospitals.

The remainder of the paper is organised as follows:
Section~\ref{sec:collection} describes the data collection process and the
ethics protocol followed to obtain the data; Section~\ref{sec:statistics}
presents corpus statistics and lexical analysis; Section~\ref{sec:annotations}
introduces the annotated subset, the CRF scheme, and the CRF Filling task;
Section~\ref{sec:experiments} describes the experimental setup and presents
zero-shot baseline results with two large language models;
Section~\ref{sec:related} surveys related work; and Section~\ref{sec:conclusion}
concludes.

\section{Data Collection and Ethics Protocol}
\label{sec:collection}

The eCREAM-MedCorpus dataset was collected under a research protocol submitted to and approved by the hospital's ethics committees (ECs). Specifically, for two Italian hospitals (San Giovanni Bosco in Turin, and Sant'Andrea hospital in Vercelli), the entire process, including execution of the data transfer agreement, has been fully completed.

This section describes the institutional approval process, the scope of the authorised data access, and the constraints imposed on data usage and redistribution, ensuring full compliance with Italian and European data protection regulations (GDPR).

\subsection{NLP Study Protocol}
Data collection was carried out in the context of the NLP-DeVal study (Development and Validation of a Natural Language Processing Tool to Enable Clinical Research in Emergency and Acute Care Medicine; ClinicalTrials.gov: NCT06240572), a retrospective, observational, multicentre study and the first pillar of the broader eCREAM European project (Horizon Europe Grant Agreement no. 101057726). The study protocol aims to exploit and validate a large language model in extracting clinically relevant information from free-text emergency department (ED) notes across five languages (English, Greek, Italian, Polish, and Slovenian). To this end, each participating hospital was asked to provide a large set of free-text clinical notes extracted from their EHR systems, covering patients who visited the ED between 2021 and 2023. Prior to any transfer, notes undergo a two-stage anonymisation process: first, all references to the patient and clinical context (names, dates, diagnoses, disposition) are removed; second, a certified anonymisation software (AnonymAI) is applied to eliminate residual references to third parties (e.g., names and contact details of relatives). The resulting anonymised texts are then transmitted to the central eCREAM server for the analyses. A subset of about 5,000 notes per language is annotated by trained emergency physicians against the eCREAM virtual Case Report Form (vCRF), and used for model fine-tuning and validation, with concordance measured via Cohen’s $\kappa$ (acceptability threshold: $\kappa$ > 0.75).

\subsection{Submission and Approval Process}
The protocol was submitted to the ECs of 26 participating centres across seven countries: 18 Italian, 3 UK, 2 Polish, 1 Slovenian, 1 Greek, and 1 Swiss. Overall, the protocol received approval from the majority of ECs. However, 7 centres did not reach a positive outcome: 4 Italian ECs raised objections that proved irresolvable within the framework of the study design, 1 Swiss EC denied authorisation, citing privacy-related concerns regarding cross-border data transfer, and approvals from 1 Polish and 1 Greek EC are still pending.
The approval process proved considerably more time-consuming than anticipated. In several cases, EC deliberation extended beyond twelve months from the date of submission. This delay appears to reflect a broader unpreparedness of ethics committees to evaluate studies of this nature (specifically, retrospective studies involving anonymised free-text EHR data and NLP-based processing) rather than substantive ethical objections. The novel combination of retrospective design, large-scale text data, and cross-institutional data sharing fell outside the frameworks most ECs were accustomed to, generating extended procedural uncertainty.

Beyond ethics approval itself, the formalisation of data transfer agreements between the coordinating centre and the individual hospitals proved equally time-demanding. Institutional legal and data protection offices frequently requested detailed clarifications on privacy safeguards, data flows, and responsibilities under GDPR before co-signing the agreements. This further extended the onboarding timeline for several centres, in some cases adding months to an already protracted process.

Taken together, these observations highlight a structural gap in the current regulatory landscape: while GDPR provides a robust framework for data protection, the practical implementation of ethics review for NLP-based, retrospective studies remains insufficiently standardised across European institutions, representing a significant bottleneck for this type of research.



\section{eCREAM-MedCorpus: Corpus Statistics}
\label{sec:statistics}

\subsection{Overall Corpus}
The current version of eCREAM-MedCorpus is composed by clinical notes coming from the Emergency Department of two Italian hospitals, the San Giovanni Bosco hospital in Turin (SGB), and the Sant'Andrea hospital in Vercelli  (SA).
The notes have been collected among a three-year span, 2021-2023, and they represent all free-text documents produced by the Emergency Departments of the two hospitals. Each note is assigned to a different category based on the context it is written in. There are $10$ categories in common between the two datasets, i.e. Anamnesis, Clinical Diary, Discharge, Home Based Therapy, LIS (Laboratory Information System), Medical Visit, Nursing Care Home, RIS (Radiology Information System), Specialist Consultancy, and Triage. 
For the case of the SGB dataset, there is an extra category Other Test, while for the SA is Vital Parameters.

The full corpus comprises approximately $4.2$ millions Italian clinical notes ($1.9M$ from the SGB dataset, $2.3M$ from the SA dataset), for an overall of 221 millions words. Table~\ref{tab:stats} reports key statistics including total note count, token count, average and median note length, and note-length distribution across note types and hospitals. 
The SA dataset contains a substantial number of very short clinical notes, reflecting the practices of that specific emergency department setting. We chose to retain all notes, regardless of length, and leave any length-based filtering decisions to end users according to their specific research needs.
Figure \ref{fig:length_distribution} from Appendix \ref{app:length_distribution} reports the distribution of note lengths across categories.

\begin{table}[t]
\small
\begin{tabular}{l|l|rrrrr}
\toprule
\textbf{Note category} & \textbf{Dataset} & \textbf{N notes} & \textbf{N words} & \textbf{Mean} & \textbf{Median} & \textbf{Stdev} \\
\toprule
ANAMNESIS & SGB & 179.098 & 10.907.482 & 60.9 & 43.0 & 56.2 \\
 & SA & 143.864 & 1.390.299 & 9.7 & 4.0 & 14.2 \\
\midrule
CLINICAL DIARY & SGB & 327.182 & 14.090.953 & 43.1 & 23.0 & 55.1 \\
 & SA & 133.217 & 3.029.685 & 22.7 & 14.0 & 27.1 \\
\midrule
DISCHARGE & SGB & 246.583 & 2.091.052 & 8.5 & 7.0 & 7.7 \\
 & SA & 278.226 & 3.165.742 & 11.4 & 1.0 & 18.5 \\
\midrule
HOME BASED THERAPY & SGB & 10.877 & 217.258 & 20.0 & 13.0 & 19.0 \\
 & SA & 143.854 & 874.182 & 6.1 & 2.0 & 10.6 \\
\midrule
LIS & SGB & 174.988 & 54.512.409 & 311.5 & 367.0 & 201.4 \\
 & SA & 176.346 & 45.583.945 & 258.5 & 285.0 & 150.0 \\
\midrule
MEDICAL VISIT & SGB & 170.240 & 6.930.859 & 40.7 & 36.0 & 25.9 \\
 & SA & 451.773 & 8.948.659 & 19.8 & 11.0 & 23.8 \\
\midrule
NURSING CARE NOTES & SGB & 325.243 & 9.720.279 & 29.9 & 20.0 & 29.0 \\
 & SA & 330.543 & 5.224.545 & 15.8 & 11.0 & 15.5 \\
\midrule
RIS & SGB & 123.126 & 16.191.912 & 131.5 & 109.0 & 68.4 \\
 & SA & 95.306 & 17.013.326 & 178.5 & 163.0 & 61.9 \\
\midrule
SPECIALIST CONSULTANCY & SGB & 38.464 & 4.555.156 & 118.4 & 98.0 & 82.1 \\
 & SA & 67.215 & 5.487.211 & 81.6 & 65.0 & 81.4 \\
\midrule
TRIAGE & SGB & 374.999 & 6.213.487 & 16.6 & 10.0 & 10.7 \\
 & SA & 180.622 & 3.054.051 & 16.9 & 11.0 & 15.8 \\\midrule
OTHER TEST & SGB & 1.454 & 312.567 & 215.0 & 202.0 & 78.6 \\
VITAL PARAMETERS & SA & 287.031 & 1.716.962 & 6.0 & 6.0 & 3.0 \\\midrule
\textbf{TOTAL} & SGB & 1.972.254 & 125.743.414 & 63.8 & 24.0 & 109.4 \\
 & SA & 2.287.997 & 95.488.607 & 41.7 & 9.0 & 86.5 \\\midrule
\textbf{TOTAL} & \textbf{BOTH} & 4.260.251 & 221.232.021 & 51.9 & 14.0 & 98.4 \\
\bottomrule
\end{tabular}
\caption{\textbf{Statistics of eCREAM-MedCorpus}. For each note category, we report SA and SGB statistics separately.}
\label{tab:stats}
\end{table}

\begin{figure}
    \centering
      \includegraphics[width=0.48\linewidth]{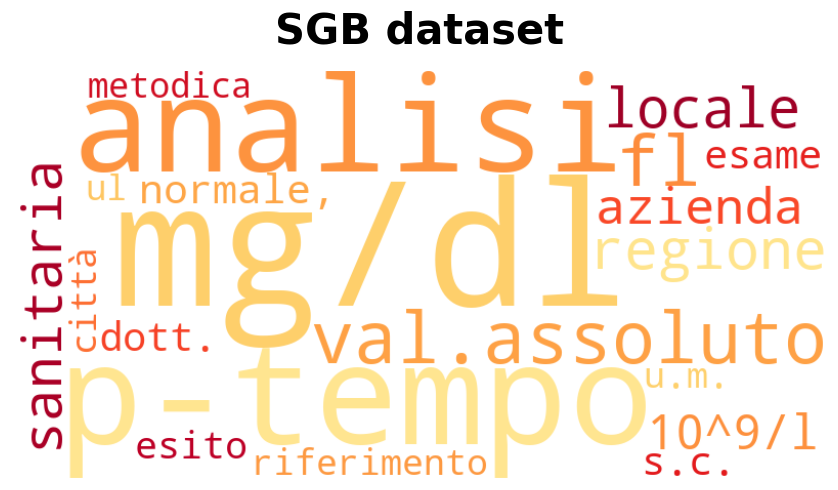} \hfill
      \includegraphics[width=0.48\linewidth]{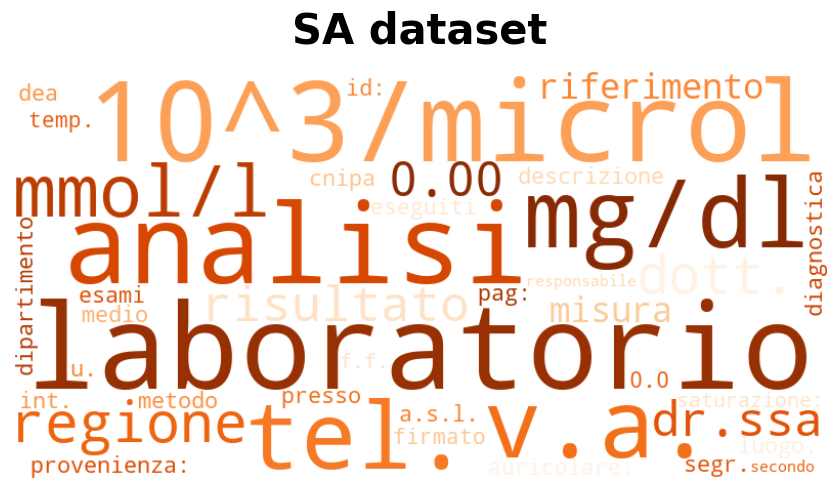} \hfill
      \includegraphics[width=0.48\linewidth]{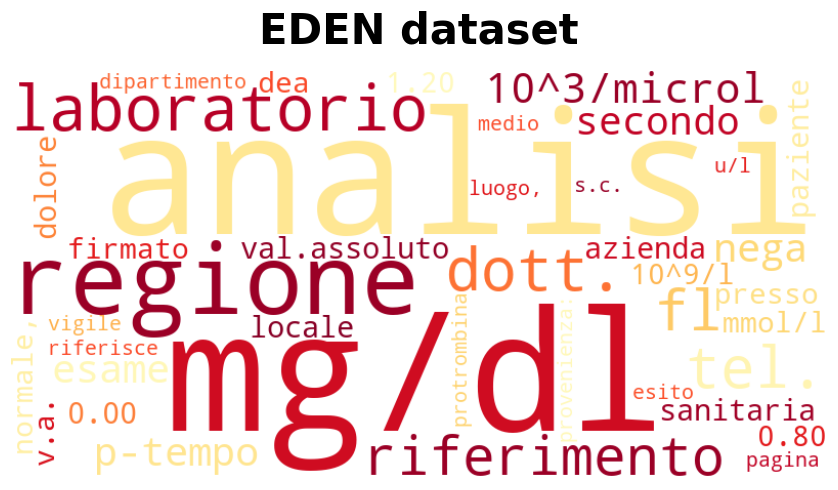} \hfill
      \includegraphics[width=0.48\linewidth]{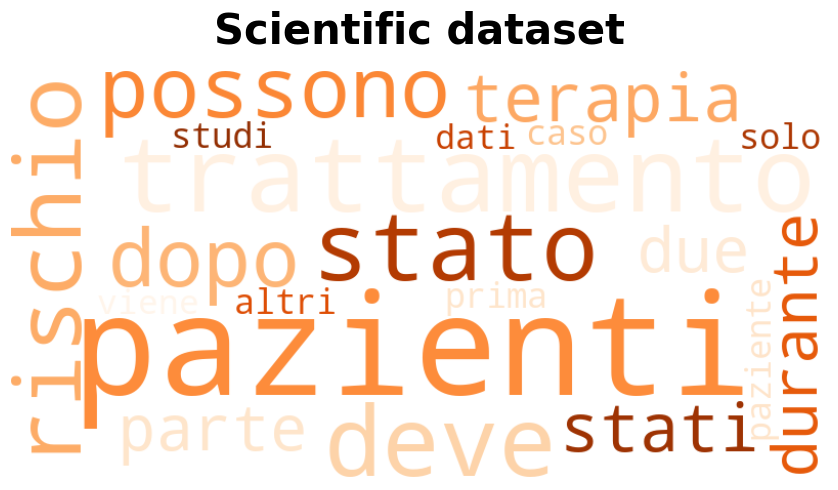} \hfill
    \caption{Wordclouds for the two sources of eCREAM-MedCorpus (SGB dataset, top-left, and VH, top-right), and the combination of the two (bottom-left). We compare them with a collection of existing resources (\textsc{Scientific Dataset}, bottom-right).}
    \label{fig:wordcloud}
\end{figure}

\subsection{Comparing eCREAM-MedCorpus with Scientific Literature in Medicine} 

In order to understand the specific characteristics of eCREAM-MedCorpus, we compare it with the collection of Italian sources in the medical domain presented by \citet{ferrazzi-etal-2026-small} (\textsc{Scientific Dataset}), including a large collection of scientific literature in medicine, drug usage instructions, Wikipedia pages about medicine, university theses in medicine-related topics, supplement descriptions, and websites on the medical domain.
Through a comparison with this resource (which is composed of most of the publicly available medical data in Italian) we seek to identify and highlight the distinctive properties of eCREAM-MedCorpus. Rather than limiting the discussion to differences in data sources and collection methodologies, we quantitatively assess a range of dataset characteristics to provide a more comprehensive comparison.

The \textsc{Scientific Dataset} comprises $280M$ words, while  eCREAM-MedCorpus  is limited to $221M$. To enforce fair comparison, we randomly sample from \textsc{Scientific Dataset} in order to reduce it to a comparable size. 

To analyse the lexical diversity of the corpus, we compute the Type-to-Token Ratio (TTR):

\begin{equation} 
\text{TTR($x$)}=\frac{\text{Number of unique tokens in $x$}}{\text{Total number of tokens in $x$}}
\end{equation}

where $x$ correspond to the whole corpus.
We observe that the TTR is higher for the \textsc{Scientific Dataset} ($2\%$) than for eCREAM-MedCorpus ($0.6\%$). This result suggests that eCREAM-MedCorpus exhibits a higher degree of lexical repetition, which is consistent with the nature of clinical documentation, where healthcare professionals commonly rely on concise formulations, recurring abbreviations, and standardized or formulaic expressions, without much room for elaborated linguistic formulations. In contrast, scientific texts tend to display a richer and more varied vocabulary, reflecting greater stylistic and topical diversity across publications.
These considerations can be derived from the number of distinct words too ($4.0M$ for \textsc{Scientific Dataset}, and $1.2M$ for eCREAM-MedCorpus).

While TTR provides a corpus-level measure of lexical diversity, it does not capture variation at the document level. To better characterize such properties within individual notes, we compute the average lexical diversity across documents. Because the notes in eCREAM-MedCorpus are substantially shorter than the texts contained in the \textsc{Scientific Dataset}, direct comparisons based on raw TTR would be biased by document length. To mitigate this effect, we compute the Corrected Type-to-Token Ratio (CTTR, \citet{carroll1964language}) for each document individually:

\begin{equation}
    \text{CTTR($x$)}=\frac{\text{Number of unique tokens in $x$}}{\sqrt{2\times\text{Total number of tokens in $x$}}}
\end{equation}

We observe that (opposite to what holds at a corpus level), eCREAM-MedCorpus notes are more dense than the \textsc{Scientific Dataset}, with a CTTR($x$) \% of $91.3$ for the former, and $82.7$ for the latter, suggesting each note in eCREAM-MedCorpus has higher degrees of lexical richness than documents and chunks taken from existing medical sources. 

Taken together, these findings highlight the specialized nature of the clinical language represented in eCREAM-MedCorpus. While the corpus as a whole exhibits a relatively repetitive vocabulary, reflecting the widespread use of standardized terminology and recurring expressions across notes, individual documents tend to be lexically dense and contain a high proportion of unique terms. This suggests that clinical notes share a common linguistic register yet convey substantial information within a limited amount of text, resulting in higher within-document lexical diversity than that observed in existing sources.
Based on these findings, we perform a qualitative analysis of the most common words in the different datasets. We summarize the results in the word clouds presented in Figure~\ref{fig:wordcloud}, where we report the $20$ most frequent words after stop words removal for the SGB split, the SA split, the composition of the two, and the \textsc{Scientific Dataset} dataset. 
While the most frequent words in eCREAM-MedCorpus reflect direct clinical
measurement and action (\textit{``analysis''}, \textit{``mg/dl''},
\textit{``absolute value''}, \textit{``doctor''}), those in existing corpora
tend to be more general and narrative in nature (\textit{``patients''},
\textit{``during''}, \textit{``risk''}, \textit{``can''}, \textit{``part''}).


\section{eCREAM-MedCorpus Annotated Subset}
\label{sec:annotations}

The goal of Use Case 1 of the eCREAM project is to evaluate the propensity of the participating EDs to hospitalise their patients. Toward this global objective, a specific interest is about patients presenting following loss of consciousness or with dyspnea at ED arrival, to evaluate whether the extracted information from the free text sections of the EHRs adds relevant predictive information to the characteristics that can be obtained from structured (not free text) sources.

In the current eCREAM-MedCorpus version, a subset of 5{,}746 clinical notes from
the SGB hospital partition has been manually annotated for training and evaluating models
on the CRF-filling task for two patient cases, loss of consciousness (CRF-1) and  dyspnea (CRF-2). Within this section we describe the task, the annotation procedure, and the resulting dataset.

\begin{table}[t]
  \caption{Distribution of CRF items by value type.}
  \label{tab:crf_categories}
  \begin{tabular}{lcc}
    \toprule
    Value Type & \# Items & \% \\
    \midrule
    Binary (Y/N)       & 87 & 65.9\% \\
    Category        & 16 & 12.1\% \\
    Number        & 20 & 15.0\% \\
    Number-or-category              & 9 & 6.8\% \\
    \midrule
    Total              & 132 & 100\% \\
    \bottomrule
  \end{tabular}
\end{table}

\subsection{CRF-filling Annotation}

Case Report Forms (CRFs) are essential tools in clinical research, designed to
systematically and consistently collect patient data across a predefined set of
items. By standardising data collection, they ensure accuracy, reliability, and
validity, properties that are crucial for producing meaningful and reproducible
results in clinical studies.
The CRF used for the eCREAM-MedCorpus collection contains 132 items organised into clinical domains.
Each item of the CRF admits a \emph{value type} that constrains the set of
valid predictions for that specific item. In the current eCREAM-MedCorpus CRF there are four value types:

\begin{itemize}
  \item \textbf{Y/N} - binary items whose valid values are \emph{Y} (yes) or
        \emph{N} (no); for instance, the item \textsc{Administration of diuretics} is constrained to \textsc{yes} or \textsc{no}.
  \item \textbf{Category} - items whose valid values form a fixed, finite set
        of categories; for instance, the item \textsc{Cardiovascular failure} is constrained to \textsc{acute} or \textsc{chronic}.
  \item \textbf{Number} - items expecting an integer or floating-point
        numerical value; for instance, the item \textsc{pH} is constrained to a numerical value.
  \item \textbf{Number-or-Category} - items whose value is a number when
        explicitly stated in the text, otherwise a category from a fixed set; for instance, the item \textsc{Heart Rate} is constrained to a numerical value, or a value between \textsc{bradycardic}, \textsc{normocardic}, or \textsc{tachycardic}.
\end{itemize}

In addition, any item of any type can receive the value \emph{unknown} when
there is insufficient evidence in the clinical note to support any valid value.
Table~\ref{tab:crf_categories} summarises the distribution of items across the admitted value 
types in the SGB partion of eCREAM-MedCorpus.

\begin{figure}[t]
\centering
\includegraphics[width=\linewidth]{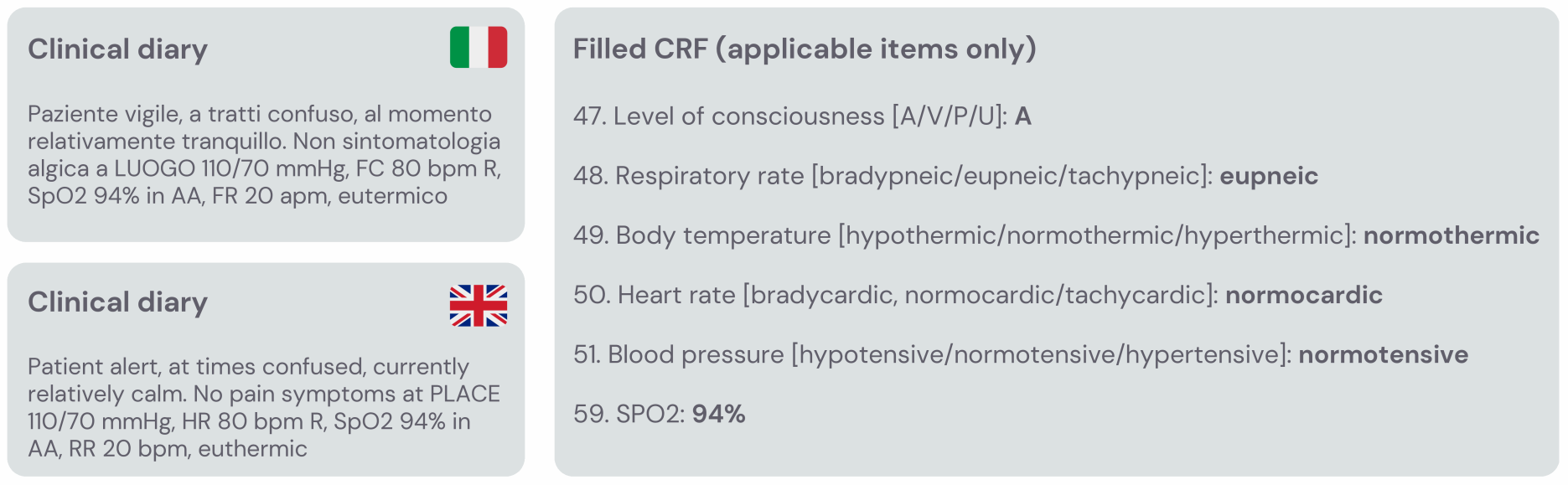}
\caption{Example of a clinical note, its English translation (provided only for reader comprehension), and the corresponding CRF items with assigned values. All other CRF fields default to \textit{unknown}.}
\label{fig:crf-example}
\end{figure}

Finally, Figure \ref{fig:crf-example} provides an example of a clinical note and the corresponding CRF annotations. Here, item 47 of the CRF, \textsc{Level of consciousness}, admits four possible categorial values, A (Alert), V (Verbal), P (Pain) and  U (Unresponsive), and the annotator of the clinical note in the Figure has selected the A value, because the note reports \textit{paziente vigile}. Item 59 of the CRF, \textsc{SPO2}, on the other side, is restricted to numerical values, and the value for the clinical note in the example is 94\%. 

\subsection{Annotation Procedure}
Given a clinical note, like the one in Figure \ref{fig:crf-example}, the task for the annotator is to identify those items in the CRF that are mentioned in the note. When an item is  identified, it is selected in a dedicated menu of the annotation tool as well as the associated value, and the portion of text supporting the item value is selected in the note.

A subset of 5{,}746 notes from the eCREAM-MedCorpus SGB partition was selected for manual annotation following criteria
designed to ensure representativeness across emergency departments and note types.
Annotation was carried out by a team of trained clinicians, each of whom
read the note in full and completed the structured CRF by filling in the value
of each item mentioned in the note, or leaving it as \emph{unknown} when no evidence was present. A
calibration phase preceded the main annotation effort, and several CRF items
whose formulation proved ambiguous were revised through consensus adjudication
before the final annotation pass.
Annotation was performed by clinicians using the Label Studio software. Each entry records the character offsets of the supporting text in the note, the corresponding text fragment, and the CRF item together with its assigned value.\footnote{The inter annotator agreement among annotators is currently under development, and will be available in the final version of the paper, if accepted for publication.}

\subsection{eCREAM-MedCorpus CRF-filling Annotated Dataset}

The annotated dataset contains 5{,}746 notes, for a total of 15{,}528 annotations (each one specifying the value of one of the 132 items). On average, each note contains 2.7 annotated items, though distribution is uneven: some are densely annotated, while about 30\% (1{,}699/5{,}746) are not filled at all. Label density generally remains modest, rarely exceeding 5 items per note, with a few cases reaching up to 25. Given that the CRF comprises 132 fields, this means that for any given note, well over 120 items on average are assigned the value \textsc{unknown}.

At the group level, the distribution of annotations is highly uneven
(Table~\ref{tab:item_distribution}). Clinical examination (25 items, 7{,}143
labels across 3{,}083 notes) and history taking (44 items, 3{,}326 labels
across 1{,}333 notes) are by far the most represented groups, together
accounting for about 67\% of all observed labels. Lab test results (23 items,
2{,}634 labels) have the highest mean label density per note
(3.63), reflecting the fact that when laboratory values are recorded they tend
to appear in clusters. Imaging test results (10 items, 1{,}062 labels),
treatment (7 items, 606 labels), diagnosis (20 items, 505 labels), and other
diagnostic test results (3 items, 252 labels) contribute progressively fewer
annotations. Across all groups, median labels per note is 1 or 2, with the
maximum reaching 17 for lab test results, indicating that while most notes
contribute a handful of annotations per group, a small number of highly
informative notes drive the tail of the distribution.

This skew reflects the clinical reality of the Emergency Department, where vital
signs and patient history are systematically recorded for nearly all patients,
while detailed laboratory findings, treatment actions, or final diagnoses are
documented only in a subset of cases, and often in dedicated note types.
Consistently, there are correlations between item groups and note categories
that follow naturally from each note's function: anamnesis notes align most
closely with history taking items, discharge notes with diagnosis items, and LIS
notes with lab test results, with analogous patterns holding for other
category-group pairs.

\begin{table}[t]
\centering
\begin{tabular}{lrrrrrr}
\toprule
Group & \makecell[r]{CRF\\items} & Notes & \makecell[r]{Total\\labels} & \makecell[r]{Mean\\labels/note} & \makecell[r]{Median\\labels/note} & \makecell[r]{Max\\labels/note} \\
\midrule
Clinical examination          & 25 & 3083 & 7143 & 2.32 & 2.0 & 9  \\
History taking                & 44 & 1333 & 3326 & 2.50 & 2.0 & 11 \\
Lab test results              & 23 & 725  & 2634 & 3.63 & 2.0 & 17 \\
Imaging test results          & 10 & 759  & 1062 & 1.40 & 1.0 & 4  \\
Treatment                     & 7  & 469  & 606  & 1.29 & 1.0 & 4  \\
Diagnosis                     & 20 & 430  & 505  & 1.17 & 1.0 & 4  \\
Other diag.\ test results     & 3  & 247  & 252  & 1.02 & 1.0 & 2  \\
\bottomrule
\end{tabular}
\caption{Distribution of annotated items per CRF group across eCREAM-MedCorpus notes.}
\label{tab:item_distribution}
\end{table}


\section{Experimental Setting}
\label{sec:experiments}

\paragraph{Dataset Splits.}
The 5{,}746 annotated notes are divided into training, development, and test
splits (70\% / 15\% / 15\%) at the note level to prevent data leakage.
The split procedure has two objectives: ensuring that every CRF item with at
least five occurrences in the corpus appears in both the development and test
sets, and preserving the distribution of note categories across splits.
To achieve the first objective, for each qualifying item one note containing
that item is reserved for the development set and one for the test set before
any further allocation. The remaining notes are then distributed using
category-stratified sampling, and all residual notes are assigned to training.
The resulting splits contain 4{,}030 training notes, 857 development notes,
and 859 test notes.

\paragraph{CRF Filling Task.}
The CRF Filling task takes as input a given  clinical note, and a
model must predict the value of each of the 132 CRF items according to its
type, or assign \emph{unknown} when no sufficient evidence is present. The task
combines structured information extraction with multi-class classification over
a heterogeneous set of output types. The item type and the list of valid values
are provided to the model as part of the prompt, since item names alone are
often insufficient to determine the expected response format.

\paragraph{Data and Models.}
All experiments are conducted on the annotated subset from the SGB hospital
described in Section~\ref{sec:annotations}. Because the experiments presented
here are preliminary (we plan to extend the study with few-shot and
fine-tuning experiments in future work) we deliberately reserve the test set
for the final evaluation stage and report all results on the \textbf{development
set} (857 notes), ensuring that test-set performance remains an unbiased
estimate for future comparisons.

We evaluate two open large language models in a zero-shot setting: Gemma-3 27B~\cite{GemmaTeam2024}, and MedGemma-27B ~\cite{MedGemma2025}, a biomedically adapted variant of Gemma-3. 
The two models form a controlled comparison: since
MedGemma-27B is derived directly from Gemma-27B, any performance difference can
be attributed to biomedical domain adaptation rather than architectural or scale
differences. All experiments were run on two NVIDIA L40S GPUs in parallel. This
setup is representative of institutional HPC infrastructure and confirms that
27B-scale models are practically deployable in such environments while remaining
competitive with larger proprietary models on medical benchmarks. Further details
on the models are provided in Appendix~\ref{app:models}.

\paragraph{Model Prompting.}
A key design choice is how to structure the interaction between the model and
the 132 CRF items. We evaluate three configurations spanning the
efficiency-controllability trade-off:
\begin{itemize}
\item Single-Item Prompting (\textsc{Item}). One prompt per item per
note (132 inference calls per note). Maximises focus on each individual item.
\item Group Prompting (\textsc{Group}). One prompt per clinical group
per note. Reduces inference calls while keeping each prompt thematically
coherent.
\item Full-Note Prompting (\textsc{Full}). All 132 items in a single
prompt per note. Most efficient, but places the greatest demand on the model's
instruction-following and output-structuring capabilities.
\end{itemize}
Full prompts for all three configurations are provided in
Appendix~\ref{app:prompts}. We additionally investigate the effect of
augmenting these base configurations with item descriptions and cautious
abstention instructions; these experiments and their results are reported in
Appendix~\ref{app:ablations}.



\paragraph{Baselines and Evaluation Metrics.}
We use two baselines: (i) Random (\textsc{Rand}). Each item is assigned a value drawn
uniformly at random from its valid value set. For numerical items, the range is
estimated from the training set and a value is sampled uniformly within
$[\min, \max]$. (ii) Most-Common (\textsc{MostCommon}). Each item is assigned its most
frequent training-set value. Since \emph{unknown} dominates every item, this
baseline always predicts \emph{unknown}, serving as a reference point for
macro-F1.

We adopt macro-averaged F1 as our
primary metric, computing the unweighted mean of per-class F1 scores. This
treats each class equally regardless of frequency, penalises models that
default to predicting \emph{unknown}, and supports meaningful comparisons
across items and datasets with different class distributions.
 We also report micro-averaged F1, weighted-F1,
and accuracy for completeness and to facilitate comparison with work using
different conventions.

\paragraph{Results and Discussion.}
Table~\ref{tab:results_main} reports all metrics for the three base
configurations evaluated with both models, plus the two baselines.
Bold values indicate the best score in each metric column.
\emph{Total predictions} reflects the total number of model queries:
\textsc{Item} generates one prediction per item per note
($132 \times 857 = 113{,}124$), \textsc{Group} one per group per note
($7 \times 857 = 5{,}999$, evaluated against all items in each group),
and \textsc{Full} one per note ($857$), later decomposed into per-item scores.

The most striking feature of Table~\ref{tab:results_main} is the wide gap
between Macro-F1 and all other reported metrics. Micro-F1, weighted-F1, and
accuracy for \textsc{MostCommon} already reach 0.978, 0.967, and 0.978 respectively, reflecting the
extreme class imbalance. The neural models also score high on these aggregate
metrics, but the margin over the trivial baseline is small and would not
meaningfully distinguish a useful extraction system from a universal abstainer.
Macro-F1, by contrast, assigns equal weight to every class including the rare
informative ones, and jumps from 0.40 (\textsc{MostCommon}) to up to 0.70 for
the best configuration, confirming it as the appropriate primary metric for this
task.

MedGemma-27B outperforms Gemma-27B on Macro-F1 in both \textsc{Item} and
\textsc{Group} configurations, confirming that biomedical domain pre-training
provides a genuine advantage for clinical information extraction in Italian. The
sole exception is \textsc{Full}, where the gap narrows and slightly reverses.
A plausible explanation is that producing 132 predictions in a single response
places heavy demand on instruction following and structured output generation, a
regime where the general instruction-tuned variant may compensate for its
reduced domain knowledge.

Moving from \textsc{Item} to \textsc{Group} to \textsc{Full} reduces inference
time by roughly $10\times$ and $20\times$ respectively (31h, 3h, and 1.5h for
MedGemma), while Macro-F1 remains within a narrow range. \textsc{Group}
emerges as the sweet spot: it achieves the highest Macro-F1 overall (0.702 for
MedGemma) while cutting runtime by a factor of ten relative to \textsc{Item}.
We hypothesise that grouping thematically related items provides the model with
useful contextual coherence, whereas \textsc{Full} prompting overwhelms it with
a long heterogeneous list that degrades per-item output quality. 

\begin{table}[t]
\caption{%
  Results on the SGB development set (857 notes, 132 CRF items).
  Macro-F1 is the primary metric.
  \textbf{Bold} marks the best value in each column (baselines excluded).
}
\label{tab:results_main}
\setlength{\tabcolsep}{4pt}
\begin{tabular}{llrrrr}
\toprule
\multicolumn{2}{l}{\textbf{Configuration}} &
  \textbf{Macro-F1} &
  \textbf{Micro-F1} &
  \textbf{Wgt-F1} &
  \textbf{Acc.} \\
\midrule
\multicolumn{2}{l}{\textsc{Random}}      & 0.030 & 0.044 & 0.061 & 0.045 \\
\multicolumn{2}{l}{\textsc{MostCommon}}  & 0.404 & 0.978 & 0.967 & 0.978 \\
\midrule
\multirow{2}{*}{\textsc{Item}}
  & Gemma     & 0.602 & 0.864 & 0.912 & 0.861 \\
  & MedGemma  & 0.670 & 0.937 & 0.958 & 0.937 \\
\midrule
\multirow{2}{*}{\textsc{Group}}
  & Gemma     & 0.672 & 0.947 & 0.964 & 0.947 \\
  & MedGemma  & \textbf{0.702} & \textbf{0.971} & \textbf{0.977} & \textbf{0.971} \\
\midrule
\multirow{2}{*}{\textsc{Full}}
  & Gemma     & 0.675 & 0.971 & 0.975 & 0.971 \\
  & MedGemma  & 0.669 & 0.975 & 0.977 & 0.975 \\
\bottomrule
\end{tabular}
\end{table}

\section{Related Work}
\label{sec:related}
In this section we report related work, addressing two relevant aspects, available medical data for Italian, and information extraction in the medical domain for Italian.


\paragraph{European Clinical Case Corpus - E3C.} The European Clinical Case Corpus (E3C) v2.0 \cite{Magnini2023} \cite{magnini-etal-2020-e3c} \cite{ghosh2025lowresourceinformationextractioneuropean}  is a freely available
multilingual corpus encompassing five languages, i.e. English, Italian, French, Spanish, and Basque\footnote{E3C-2.0 is released under Creative Commons License Attribution 4.0 International (CC BY 4.0) and is available for download from the project's website: 
\url{https://e3c.fbk.eu/data}}. It consists of clinical narratives manually annotated with semantic information, thus allowing for linguistic analysis, benchmarking, and training of information extraction systems. E3C consists of journal abstracts available from PubMed\footnote{\url{https://pubmed.ncbi.nlm.nih.gov}} and 
clinical cases published in journals (such as The Pan African Medical Journal) or available from medical training resources (e.g., the SPACCC: Spanish Clinical Case Corpus\footnote{\url{https://github.com/PlanTL-GOB-ES/SPACCC}}) and admission tests for specialties in medicine. Texts in the five languages have been collected independently but following the same criteria, making them comparable corpora. 



\paragraph{Data from clinical settings.}
While a number of large datasets for the medical domain have been made publicly available throughout the years (PubMed, Medical Wikipedia, Medical Common Crawl etc.), very few come from real clinical settings. 
MIMIC III \citep{Johnson2016}, 
MIMIC IV  \citep{johnson2023mimiciv},
ib2b-10 by \citep{Uzuner2011}, 
ib2b-14 \cite{Stubbs2015} 1.304 longitudinal medical records describing 296 patients.
In the case of Italian, E3C \citep{Magnini2023} represents the largest collection of clinical cases currently available. However, it is primarily composed of reports and educational material, which differ substantially from routine clinical documentation. Consequently, there remains a lack of large-scale datasets capturing authentic clinical narratives produced during everyday healthcare practice, especially in emergency care settings.
This is due to difficulties in data collection, and high privacy concerns connected to that.

\paragraph{Italian medical NLP tasks.}
Clinical information extraction (IE) is a long-standing area in NLP. Several tasks and datasets have been proposed throughout the years, mainly focusing on English.  
\citet{ferrazzi-etal-2026-small} presented a comprehensive collection of resources for Italian, which we summarize in the following, aiming to present an overview of landscape of publicly available data sources for IE. 
We identify three main tasks: Named Entity Recognition (NER), Relation Extracion (RE), and Slot Filling (SL). 
NER is the task that has received the highest degrees of attention, with several publicly available resources: E3C \citep{Magnini2023} and its projecteed version \citep{DBLP:journals/corr/abs-2503-20568}, PharmaER \cite{zugarini2025pharmaer}, CardioCCC \citep{448}, DisteMIST  \citep{416}, and PsyNIT \citet{CREMA2023104557}.
While many works have focused on RE for Italian \citep{ALICANTE2016263, VIANI2018140, attardi2014adapting}, the only publicly available source is represented by E3C \citep{Magnini2023}.

\paragraph{Case Report Form filling.}
Automating the population of CRFs from clinical narratives has emerged as an important research direction. Initial studies demonstrated the feasibility of extracting structured variables from free-text clinical reports \citep{MacKenzie2016}, with later work expanding both the scope and applicability of these approaches \citep{Gutierrez-Sacristan2024}. Nevertheless, most existing systems remain grounded in rule-based pipelines, keyword matching, and terminology-based mappings. While effective for well-defined extraction tasks, such methods struggle to capture contextual and semantic information, limiting their performance compared with modern NLP approaches based on representation learning and large language models.
The only exception is represented by \citet{10651607}, who tackle the CRF task using a BERT-based approach. Nevertheless, the authors did not release any of the data they based their approach on, due to privacy reasons.
To overcome privacy-related limitations, \citet{ferrazzi-etal-2025-converting} propose a dataset for CRF task built on top of publicly available resources.
Following, \citet{ferrazzi2026crf} presented a shared task with the objective of developing strategies to address the CRF filling task with state-of-the-art NLP approaches, including a small set of examples similar to the ones we release, highlighting the  lack of open data from real clinical settings.

\paragraph{Medicine-adapted models.}
There has been a growth of the interest in encoders adaptation to the medical domain, through different techniques. Examples are BioGPT~\citep{BioGPT2022},  MedPALM-1~\citep{Singhal2023}, Meditron \citep{chen2023meditron70bscalingmedicalpretraining}, PMC-LLaMA~\citep{Wu2024}, BioMistral \citep{labrak-etal-2024-biomistral}, MedPalm-2~\citep{Singhal2025}, MedPhi~\citep{corbeil-etal-2025-modular}, MedGemma~\citep{MedGemma2025}.
More recently, with the raise or reasoning models, several have been proposed with a focus on the medical domain. It is the case of Huatuo-o1 \citep{chen-etal-2025-towards-medical}, MedReason \citep{wu2025medreasonelicitingfactualmedical}, m1 \citep{huang2025m}, ReasonMed \citep{sun-etal-2025-reasonmed}. While these models are all English based, \citet{ferrazzi2026multilingualmedicalreasoningquestion} proposed the first medical reasoner for Italian.

\section{Conclusion}
\label{sec:conclusion}


We  presented eCREAM-MedCorpus, a large-scale Italian clinical corpus of
approximately 4 million notes sourced from two hospitals, accompanied by a
manually annotated subset of 5{,}746 notes labelled with 132 structured CRF
items. To our knowledge, this is the largest publicly released corpus of Italian
clinical text, and one of the most richly annotated clinical resources in any
language in terms of structured output diversity, spanning binary, categorical,
numerical, and mixed item types across multiple clinical domains.
We introduce the CRF Filling task as a structured information extraction benchmark
and provided the first zero-shot baseline results using Gemma-3 27B and
MedGemma-27B varying prompting
configuration and strategy. Our results show that large language models can
extract meaningful clinical information from Italian notes without any
task-specific supervision, with MedGemma-27B consistently outperforming its
general-purpose counterpart and group-level prompting emerging as the best
trade-off between extraction quality and inference efficiency. 

Several directions remain open for future work. On the modelling side, we plan
to extend the evaluation to few-shot and fine-tuning regimes, which we expect to
close a significant portion of the remaining gap, particularly for rare item
types and domain-specific numerical values. On the data side, the existing dataset and corresponding annotations will be extended and more data from other EDs will be publicly released as well. We hope
eCREAM-MedCorpus will catalyse research contribution to Italian clinical language processing, structured information extraction from medical text, and the evaluation
of LLMs in high-stakes biomedical settings where reliability and interpretability are paramount.

\section*{Declaration on Generative AI}
  
During the preparation of this work, the authors used generative AI for grammar and spelling review. The authors carefully reviewed and edited all generated suggestions as needed and take full responsibility for the content of this publication.




\bibliography{references}

\appendix

\section{Note Length Distribution by Category}
\label{app:length_distribution}

\begin{figure}[t]
  \centering
  \includegraphics[width=\linewidth]{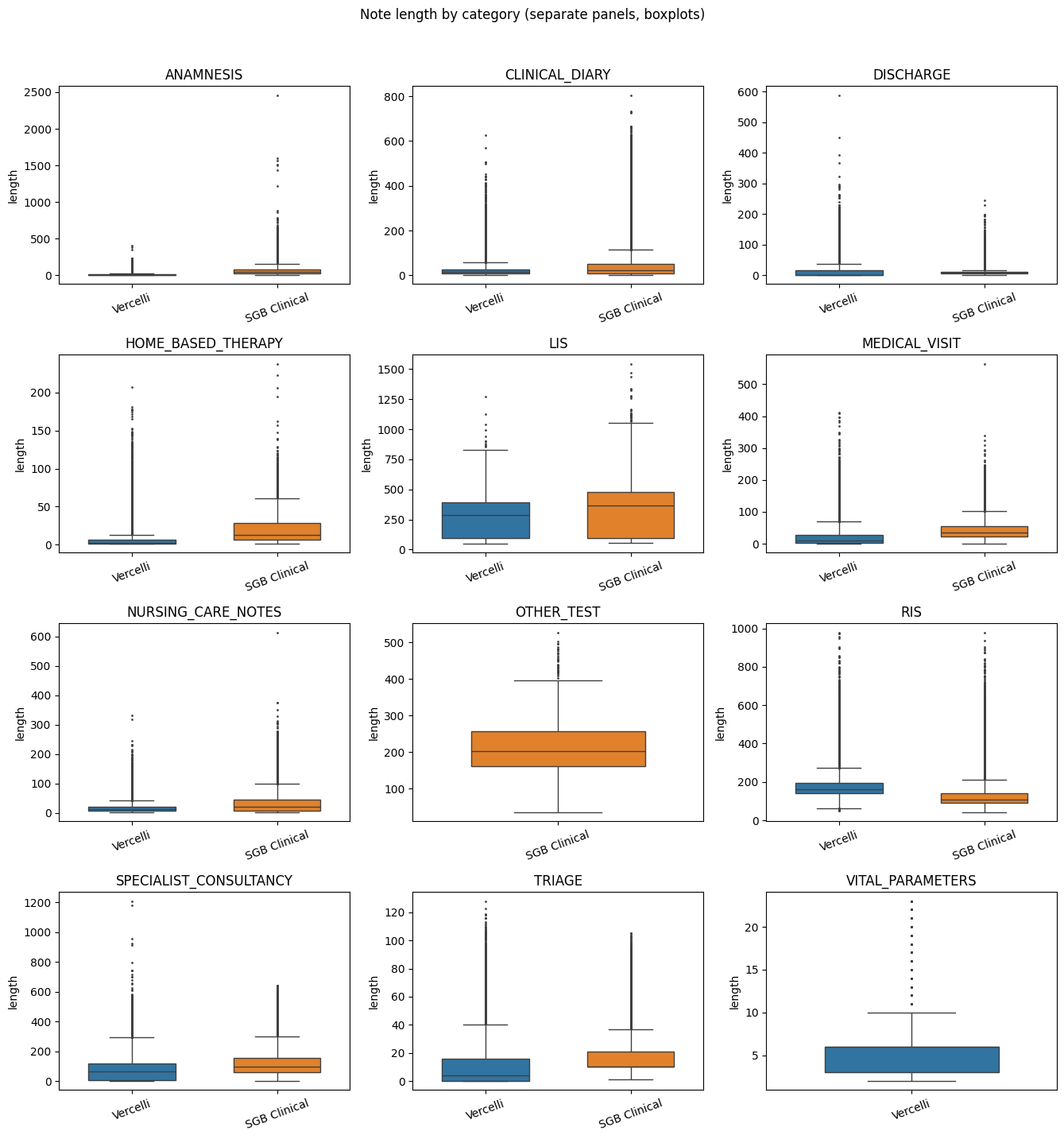}
  \caption{%
    Note length (tokens) by category and hospital source.
    Blue: Vercelli hospital (VH); orange: San Giovanni Bosco (SGB).
    Boxes show the interquartile range; whiskers extend to $1.5\times$IQR;
    points beyond the whiskers are individual outliers.
    Categories absent from one source (OTHER\_TEST, SPECIALIST\_CONSULTANCY,
    VITAL\_PARAMETERS) have no corresponding box for that source.
  }
  \label{fig:length_distribution}
\end{figure}

Figure~\ref{fig:length_distribution} shows the distribution of note lengths
(in tokens) across the 12 note categories present in the corpus, broken down
by hospital source (Vercelli, VH, in blue; San Giovanni Bosco, SGB, in orange).
Several patterns are worth noting. LIS and RIS notes are the longest on
average, reflecting the verbose, structured format of laboratory and radiology
reports. At the opposite extreme, VITAL\_PARAMETERS and TRIAGE notes are
consistently short, as they record a narrow set of numerical measurements or
brief categorical assessments. Across most categories, SGB notes
tend to be slightly longer than their SA counterparts, though the medians
remain broadly comparable. All distributions are right-skewed, with
substantial outlier mass above the upper whisker, which is consistent with the
clinical practice of appending structured annexes or repeated measurements to a
single note record.

\section{Model Details}
\label{app:models}

Table~\ref{tab:model_details} summarises the key characteristics of the two
models evaluated in this work.

\begin{table}[h!]
\centering
\begin{tabular}{lllr}
\toprule
\textbf{Model} & \textbf{Owner} & \textbf{HuggingFace ID} & \textbf{\makecell[r]{Knowledge\\cutoff}} \\
\midrule
Gemma-27B    & Google DeepMind & \texttt{google/gemma-3-27b-it}      & Aug 2024 \\
MedGemma-27B & Google DeepMind &  \texttt{google/medgemma-27b-text-it} & May 2024 \\
\bottomrule
\end{tabular}
\caption{Details of the models used in the experiments.}
\label{tab:model_details}
\end{table}

\section{Prompt Templates}
\label{app:prompts}

This appendix presents the prompt templates used for the three prompting
configurations described in Section~\ref{sec:experiments}. All prompts
share a common structure: a system role declaration, a note-type hint, a
specification of the items to extract with their expected value format, optional
modifiers (\textsc{+Desc} and \textsc{+Caut}), and the clinical note delimited
by ASCII markers. The model is always instructed to output a JSON object and
nothing else.

For all configurations, the \textbf{answer format} for each item is determined
by its value type (see Section~\ref{sec:annotations}):
\begin{itemize}
  \item \textbf{YN}: \texttt{"Y"}, \texttt{"N"}, or \texttt{"unknown"}
  \item \textbf{Number}: a numeric value (decimal separator normalised to
        \texttt{.}), or \texttt{"unknown"}
  \item \textbf{Category}: one of the item's predefined values, or
        \texttt{"unknown"}
  \item \textbf{Number-or-Category}: a numeric value or one of the predefined
        categories, or \texttt{"unknown"}
\end{itemize}

The \textsc{+Desc} strategy includes the auto-generated item description in
parentheses after the item name (e.g.\ \texttt{ITEM\_NAME (description)}).
The \textsc{+Caut} strategy appends the sentence \emph{``Be very cautious!
Unless the item is explicitly mentioned, assume that the correct value is
`unknown'.''}\ to the instructions.  Both modifications are absent in the
\textsc{Base} variant and present simultaneously in \textsc{+Desc+Caut}.

\subsection{\textsc{Item} Prompt}

One prompt is issued per CRF item per note. The model is asked to extract a
single item and return a JSON object with a single \texttt{value} key
(or multiple measurement keys for \texttt{allow\_followup} items).

\begin{lstlisting}[basicstyle=\small\ttfamily, frame=single,
                   breaklines=true, breakatwhitespace=false,
                   caption={Single-item prompt template (\textsc{Item}).
                   Fields in angle brackets are filled at runtime.},
                   label=lst:prompt_item]
You are a clinical information extraction assistant.

You will be given a clinical note from the section "<NOTE_CATEGORY>".
(domain: "<GROUP>").
Your task is to extract the value for the following clinical item:

  Item: <ITEM_NAME>

Respond with a JSON object in the following format:
{"value": <your answer>}

The value must be <VALUE_DESC>

[+Desc]  <ITEM_DESCRIPTION>

If a value is not mentioned or you are not sure, respond with "unknown".
[+Caut]  Be very cautious! Unless the item is explicitly mentioned,
         assume that the correct value is "unknown".

Do not add any explanation. Output only the JSON object.

-- CLINICAL NOTE --
<NOTE_TEXT>
-- END OF NOTE --
\end{lstlisting}

\subsection{\textsc{Group} Prompt}

One prompt is issued per clinical group per note. All items belonging to the
group are listed together with their expected format, and the model returns a
single JSON object with one key per item.

\begin{lstlisting}[basicstyle=\small\ttfamily, frame=single,
                   breaklines=true, breakatwhitespace=false,
                   caption={Group prompt template (\textsc{Group}).},
                   label=lst:prompt_group]
You are a clinical information extraction assistant.

You will be given a clinical note from the section "<NOTE_CATEGORY>".
Your task is to extract values for the following clinical items
from the domain "<GROUP>".

For each item, output a JSON object with the structure described below.
If a value is not mentioned or you are not sure, use "unknown".
[+Caut]  Be very cautious! Unless an item is explicitly mentioned,
         assume that the correct value is "unknown".

Do not add any explanation. Output only the JSON object.

-- ITEMS AND EXPECTED FORMAT --
  "<ITEM_1> [(<DESC_1>)]": {"value": <value>}  <- <VALUE_DESC_1>
  "<ITEM_2> [(<DESC_2>)]": {"value": <value>}  <- <VALUE_DESC_2>
  ...

-- EXPECTED OUTPUT FORMAT --
A single JSON object with one key per item:
{
  "<ITEM_1>": {"value": <value>},
  "<ITEM_2>": {"value": <value>},
  ...
}

-- CLINICAL NOTE --
<NOTE_TEXT>
-- END OF NOTE --
\end{lstlisting}

\subsection{\textsc{Full} Prompt}

A single prompt is issued per note, containing all 132 CRF items grouped by
clinical domain for readability. The model returns one JSON object with 132
keys.

\begin{lstlisting}[basicstyle=\small\ttfamily, frame=single,
                   breaklines=true, breakatwhitespace=false,
                   caption={Full-note prompt template (\textsc{Full}).},
                   label=lst:prompt_full]
You are a clinical information extraction assistant.

You will be given a clinical note from the section "<NOTE_CATEGORY>".
Your task is to extract values for ALL of the following clinical items.

For each item, output a JSON object with the structure described below.
If a value is not mentioned or you are not sure, use "unknown".
[+Caut]  Be very cautious! Unless an item is explicitly mentioned,
         assume that the correct value is "unknown".

Do not add any explanation. Output only the JSON object.

-- ITEMS AND EXPECTED FORMAT --
  # <GROUP_1>
  "<ITEM_1> [(<DESC_1>)]": {"value": <value>}  <- <VALUE_DESC_1>
  "<ITEM_2> [(<DESC_2>)]": {"value": <value>}  <- <VALUE_DESC_2>

  # <GROUP_2>
  "<ITEM_3> [(<DESC_3>)]": {"value": <value>}  <- <VALUE_DESC_3>
  ...

-- EXPECTED OUTPUT FORMAT --
A single JSON object with one key per item:
{
  "<ITEM_1>": {"value": <value>},
  "<ITEM_2>": {"value": <value>},
  ...
}

-- CLINICAL NOTE --
<NOTE_TEXT>
-- END OF NOTE --
\end{lstlisting}

\subsection{Item Description Generation}
\label{app:desc_prompt}

As described in Appendix~\ref{app:ablations}, the \textsc{+Desc}
strategy incorporates a brief natural-language description of each CRF item
into the prompt. These descriptions are generated in a preliminary step
by querying MedGemma-27B once per item (132 calls in total) using the
prompt shown in Listing~\ref{lst:prompt_desc}. Each description is then cached
and reused across all notes.

\begin{lstlisting}[basicstyle=\small\ttfamily, frame=single,
                   breaklines=true, breakatwhitespace=false,
                   caption={Prompt used to generate a natural-language
                   description for each CRF item (\textsc{+Desc}).},
                   label=lst:prompt_desc]
Provide a brief, medically accurate description of the clinical
item '<ITEM_NAME>', in only 1 sentence. Focus on what it is and
how it is typically documented in clinical notes, without
mentioning the specific value to extract.
\end{lstlisting}

\section{Prompt Ablations: Item Descriptions and Cautious Abstention}
\label{app:ablations}

Beyond the three base configurations evaluated in Section~\ref{sec:experiments},
we explore three additional prompt modifications applied orthogonally to each
configuration, yielding twelve conditions in total. All ablations use Gemma-3-27B and MedGemma-27B as large language models.

\paragraph{Additional modifications.}
\begin{itemize}
\item Item Description (\textsc{+Desc}). A brief natural-language description
of each CRF item (generated by MedGemma-27B in a preliminary zero-shot step,
as described in Appendix~\ref{app:desc_prompt}) is incorporated into the prompt
alongside the item name and valid values. This grounds the model's
interpretation of abbreviated or ambiguous item names.
\item Cautious Abstention (\textsc{+Caut}). The prompt is augmented with an
explicit instruction to prefer \emph{unknown} unless the target value is
unambiguously stated in the text, counteracting the model's tendency to
over-predict specific values on a dataset heavily skewed toward abstention.
\item Description + Cautious Abstention (\textsc{+Desc+Caut}). Both
modifications are applied simultaneously to test whether they are complementary
or whether one subsumes the other.
\end{itemize}

\paragraph{Results.}
Table~\ref{tab:results_full} reports all 26 conditions (24 + the two baselines). Figures~\ref{fig:macro_f1}
and~\ref{fig:runtime} visualise Macro-F1 and runtime across all configurations.

\begin{table*}[t]
\caption{%
  Full experimental results including prompt ablations on the SGB development
  set (857 notes, 132 CRF items). Macro-F1 is the primary metric.
  \textbf{Bold} marks the best value in each metric column (baselines excluded).
}
\label{tab:results_full}
\setlength{\tabcolsep}{5pt}
\begin{tabular}{llr rrrr r}
\toprule
\multicolumn{2}{l}{\textbf{Configuration}} &
  \textbf{\#Pred} &
  \textbf{Macro-F1} &
  \textbf{Micro-F1} &
  \textbf{Wgt-F1} &
  \textbf{Acc.} &
  \textbf{Runtime} \\
\midrule
\multicolumn{2}{l}{\textsc{Random}}      & 113,124 & 0.030 & 0.044 & 0.061 & 0.045 & -- \\
\multicolumn{2}{l}{\textsc{MostCommon}}  & 113,124 & 0.404 & 0.978 & 0.967 & 0.978 & -- \\
\midrule
\multirow{8}{*}{\rotatebox{90}{\textsc{Item}}}
  & \textsc{Base}~Gemma          & 113,124 & 0.602 & 0.864 & 0.912 & 0.861 & 31h\,25m \\
  & \textsc{Base}~MedGemma       & 113,124 & 0.670 & 0.937 & 0.958 & 0.937 & 31h\,15m \\
  & \textsc{+Desc}~Gemma         & 113,124 & 0.609 & 0.861 & 0.910 & 0.860 & 31h\,53m \\
  & \textsc{+Desc}~MedGemma      & 113,124 & 0.668 & 0.936 & 0.957 & 0.935 & 31h\,45m \\
  & \textsc{+Caut}~Gemma         & 113,124 & 0.611 & 0.883 & 0.924 & 0.880 & 32h\,53m \\
  & \textsc{+Caut}~MedGemma      & 113,124 & 0.679 & 0.947 & 0.964 & 0.947 & 32h\,49m \\
  & \textsc{+Desc+Caut}~Gemma    & 113,124 & 0.623 & 0.891 & 0.929 & 0.889 & 31h\,22m \\
  & \textsc{+Desc+Caut}~MedGemma & 113,124 & 0.682 & 0.948 & 0.964 & 0.947 & 31h\,34m \\
\midrule
\multirow{8}{*}{\rotatebox{90}{\textsc{Group}}}
  & \textsc{Base}~Gemma          & 5,999 & 0.672 & 0.947 & 0.964 & 0.947 & 3h\,16m \\
  & \textsc{Base}~MedGemma       & 5,999 & \textbf{0.702} & 0.971 & 0.977 & 0.971 & 3h\,1m  \\
  & \textsc{+Desc}~Gemma         & 5,999 & 0.658 & 0.922 & 0.948 & 0.921 & 4h\,2m  \\
  & \textsc{+Desc}~MedGemma      & 5,999 & 0.695 & 0.964 & 0.973 & 0.964 & 3h\,42m \\
  & \textsc{+Caut}~Gemma         & 5,999 & 0.681 & 0.959 & 0.970 & 0.959 & 3h\,19m \\
  & \textsc{+Caut}~MedGemma      & 5,999 & 0.701 & 0.973 & \textbf{0.978} & 0.973 & 3h\,6m  \\
  & \textsc{+Desc+Caut}~Gemma    & 5,999 & 0.667 & 0.941 & 0.960 & 0.941 & 4h\,4m  \\
  & \textsc{+Desc+Caut}~MedGemma & 5,999 & 0.698 & 0.972 & 0.977 & 0.972 & 3h\,48m \\
\midrule
\multirow{8}{*}{\rotatebox{90}{\textsc{Full}}}
  & \textsc{Base}~Gemma          & 857 & 0.675 & 0.971 & 0.975 & 0.971 & 1h\,37m \\
  & \textsc{Base}~MedGemma       & 857 & 0.669 & 0.975 & 0.977 & 0.975 & 1h\,25m \\
  & \textsc{+Desc}~Gemma         & 857 & 0.659 & 0.964 & 0.971 & 0.963 & 1h\,42m \\
  & \textsc{+Desc}~MedGemma      & 857 & 0.663 & 0.972 & 0.976 & 0.972 & 1h\,37m \\
  & \textsc{+Caut}~Gemma         & 857 & 0.675 & 0.972 & 0.975 & 0.972 & 1h\,35m \\
  & \textsc{+Caut}~MedGemma      & 857 & 0.671 & \textbf{0.976} & \textbf{0.978} & \textbf{0.976} & 1h\,32m \\
  & \textsc{+Desc+Caut}~Gemma    & 857 & 0.655 & 0.966 & 0.972 & 0.965 & 1h\,40m \\
  & \textsc{+Desc+Caut}~MedGemma & 857 & 0.656 & 0.974 & 0.977 & 0.974 & 1h\,44m \\
\bottomrule
\end{tabular}
\end{table*}

\paragraph{Discussion.}
The \textsc{+Caut} strategy yields consistent Macro-F1 improvements over
\textsc{Base} for all \textsc{Item} conditions (+0.009 for Gemma, +0.009 for
MedGemma), while its effect is negligible in \textsc{Group} and \textsc{Full}.
This is consistent with the hypothesis that an uncalibrated model over-predicts
specific values when queried item by item, and that the cautious instruction
partially corrects this bias. When multiple items are presented simultaneously,
the model appears already more conservative, possibly because the longer output
requirement encourages shorter, safer responses.

The \textsc{+Desc} strategy does not improve and often slightly degrades
Macro-F1 relative to \textsc{Base} across all configurations. Two possible
explanations are: (i) the auto-generated descriptions may contain inaccuracies
or inappropriate generalisations for the specific hospital protocol; and (ii)
the additional prompt length may distract the model from the clinical note.
Future work should explore expert-authored descriptions or retrieval-based
grounding as alternatives.

\textsc{+Desc+Caut} achieves the overall best Macro-F1 in the \textsc{Item}
configuration for both models (0.623 Gemma, 0.682 MedGemma), suggesting that
in the single-item regime the two modifications are complementary: descriptions
help identify what to look for, while the cautious instruction prevents
confabulation when evidence is weak. For \textsc{Group} and \textsc{Full},
however, \textsc{+Desc+Caut} generally underperforms \textsc{+Caut} alone,
reinforcing the finding that description injection is unhelpful when items are
presented in group context.

\section{Experimental Results: Visualisations}
\label{app:figures}

\begin{figure}[t]
  \centering
  \includegraphics[width=\linewidth]{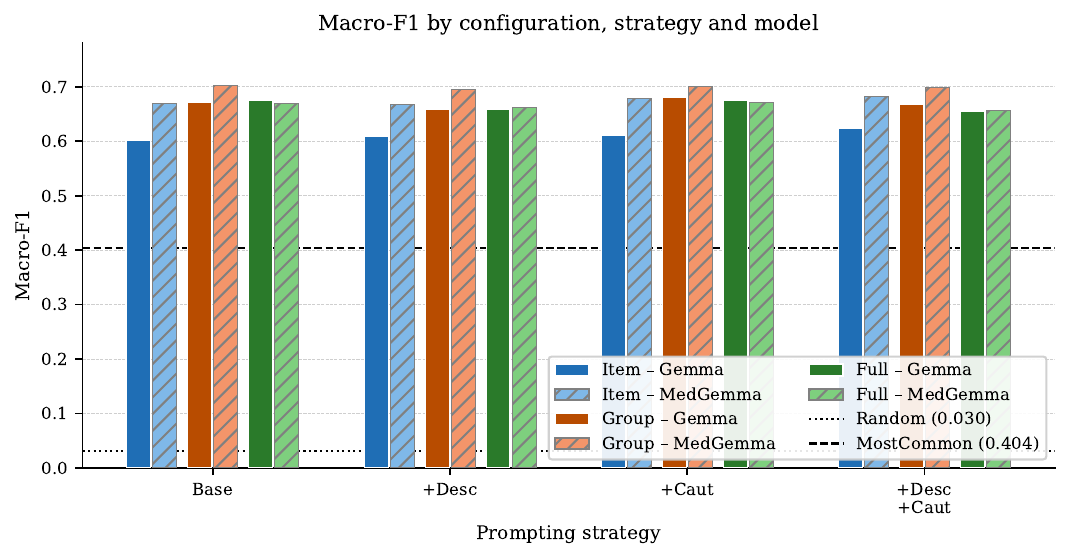}
  \caption{%
    Macro-F1 for all 24 model--strategy combinations, grouped by prompting
    configuration (\textsc{Item}, \textsc{Group}, \textsc{Full}).
    Dashed horizontal lines mark the \textsc{Random} (0.030) and
    \textsc{MostCommon} (0.404) baselines for reference.
  }
  \label{fig:macro_f1}
\end{figure}

\begin{figure}[t]
  \centering
  \includegraphics[width=\linewidth]{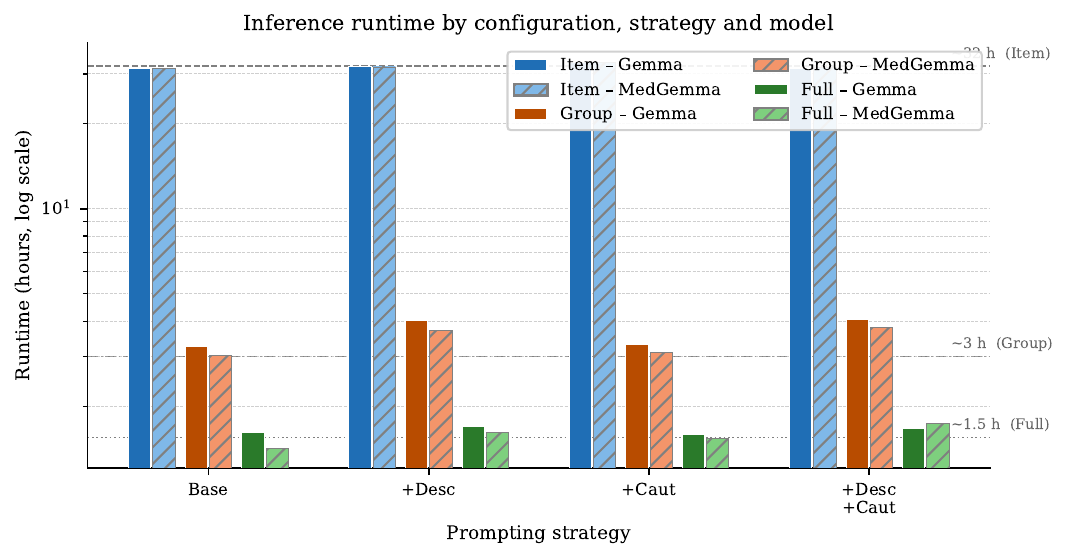}
  \caption{%
    Wall-clock inference time (logarithmic scale) for all model--strategy
    combinations. \textsc{Item}-level prompting requires roughly $10\times$
    more time than \textsc{Group} and roughly $20\times$ more than
    \textsc{Full}.
  }
  \label{fig:runtime}
\end{figure}

Figures~\ref{fig:macro_f1} and~\ref{fig:runtime} provide a visual summary of
the results discussed in Section~\ref{sec:experiments}. Figure~\ref{fig:macro_f1}
plots Macro-F1 across all 24 model-strategy combinations, making the relative
ordering of configurations and the effect of each prompting strategy immediately
readable. Figure~\ref{fig:runtime} shows the corresponding wall-clock inference
times on a logarithmic scale, illustrating the order-of-magnitude differences
between the \textsc{Item}, \textsc{Group}, and \textsc{Full} configurations.

\end{document}